\documentclass{article}
\usepackage{spconf,amsmath,graphicx}

\DeclareMathOperator*{\argmax}{argmax}

\title{Noisy Label Refinement with Semantically Reliable Synthetic Images}
\name{Yingxuan Li$^1$, Jiafeng Mao$^2$, Yusuke Matsui$^1$}
\address{$^1$The University of Tokyo, $^2$CyberAgent, Inc.}

\usepackage{xcolor}

\usepackage{algorithm}
\usepackage{algorithmic}
\usepackage{amsmath}

\usepackage{booktabs}
\usepackage{multirow} 
\usepackage{subcaption}
\usepackage{makecell}
\usepackage{tabularx}
\usepackage{enumitem}
\usepackage{balance}
\usepackage{fancyhdr}
\usepackage{url}

\begin{document}
%
\maketitle
\begin{abstract}
Semantic noise in image classification datasets, where visually similar categories are frequently mislabeled, poses a significant challenge to conventional supervised learning approaches. In this paper, we explore the potential of using synthetic images generated by advanced text-to-image models to address this issue. Although these high-quality synthetic images come with reliable labels, their direct application in training is limited by domain gaps and diversity constraints. Unlike conventional approaches, we propose a novel method that leverages synthetic images as reliable reference points to identify and correct mislabeled samples in noisy datasets \footnote{The code for the main experiments is available at: \url{https://github.com/liyingxuan1012/NoisyLabelRefinement-Syn}.}. Extensive experiments across multiple benchmark datasets show that our approach significantly improves classification accuracy under various noise conditions, especially in challenging scenarios with semantic label noise. Additionally, since our method is orthogonal to existing noise-robust learning techniques, when combined with state-of-the-art noise-robust training methods, it achieves superior performance, improving accuracy by 30\% on CIFAR-10 and by 11\% on CIFAR-100 under 70\% semantic noise, and by 24\% on ImageNet-100 under real-world noise conditions.
\end{abstract}
%
%

\section{Introduction}
\label{sec:intro}

Learning from noisy data remains one of the most fundamental and challenging problems in machine learning. This challenge is particularly pronounced in image classification, where datasets often contain images that are mislabeled with incorrect categories. 
Such mislabeling poses a significant obstacle to conventional supervised learning approaches, as incorrect labels generate misleading gradients during model training, ultimately compromising the performance of classification models. 
Although traditional research on learning from noisy labels has typically assumed that label corruption follows an independent and identically distributed (i.i.d.) pattern, where images from each category have an equal probability of being randomly mislabeled, real-world scenarios present a more complex challenge~\cite{zhang2021learning}. 
In practice, noise patterns are often feature-dependent, meaning that visually similar categories are more likely to be confused and mislabeled as each other. This semantic relationship in label noise significantly limits the effectiveness of traditional approaches that rely solely on the model's inherent generalization capability. 
Furthermore, when the dataset's inherent information becomes fundamentally compromised due to such semantic noise, conventional methods struggle to maintain reliable performance. 
To address this challenge, we propose incorporating reliable external insights to support the training of high-performance models under semantically biased noise conditions.

\begin{figure}
\centering
    \subfloat[Random noise]{
       \includegraphics[width=.23\textwidth]{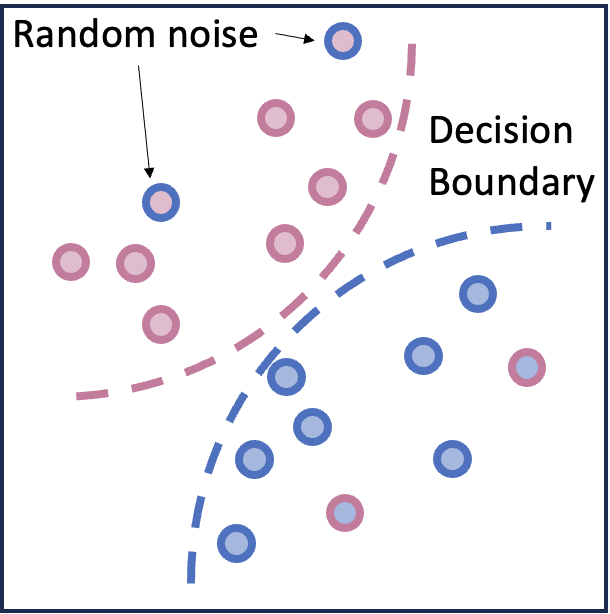}}
    \label{1a}
    \subfloat[Hard (semantic) noise]{
        \includegraphics[width=.23\textwidth]{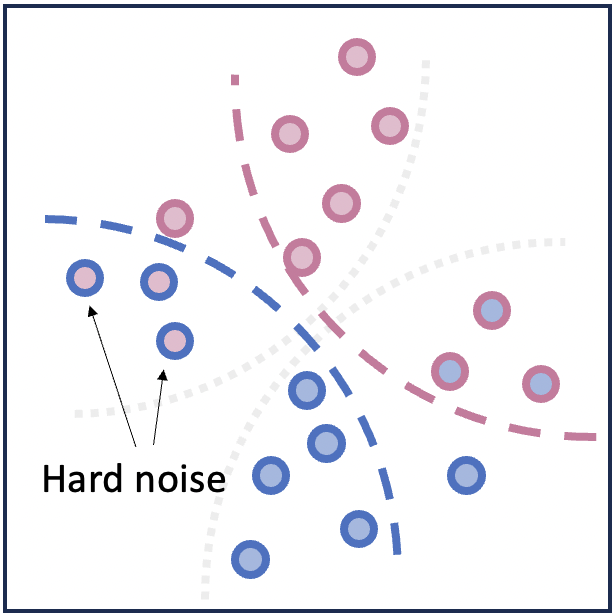}}
    \label{1b}
    \subfloat[Leverage synthetic images as stable anchors]{
        \includegraphics[width=.475\textwidth]{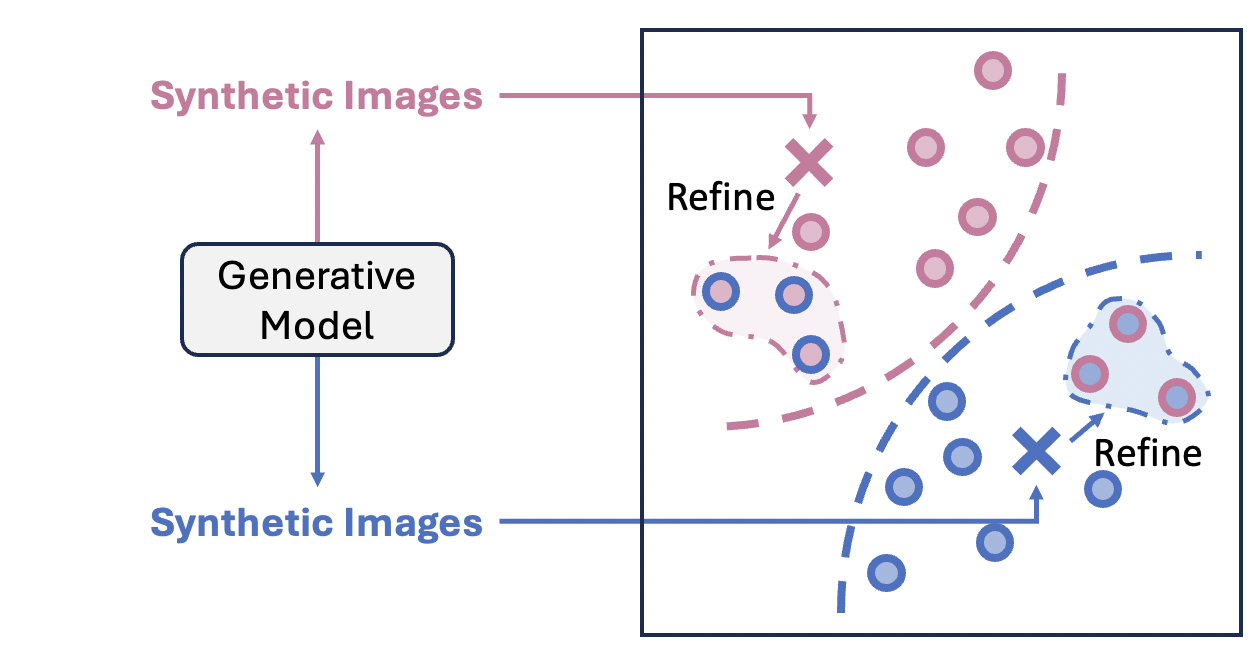}}
    \label{1c}
  \caption{Although conventional supervised learning approaches are naturally robust against random noise, they struggle with semantic noise. We propose using synthetic images as reliable references to refine these noisy labels.}
  \label{fig:ent}
\end{figure}

Recent years have witnessed the rapid emergence and development of deep generative models capable of producing high-quality realistic images. 
Text-to-image models, such as DALL-E 2~\cite{ramesh2022hierarchical} and Stable Diffusion~\cite{rombach2022high}, have demonstrated remarkable capabilities in generating photorealistic images from simple textual descriptions. 
A crucial characteristic of these synthetic images is that they inherently carry accurate labels through their generating text prompt. 
While this feature makes synthetic images appealing for machine learning datasets, their direct application in standard image classification training has shown limited success and can sometimes lead to degraded model performance, primarily due to constrained diversity and domain gaps between synthetic and target datasets~\cite{hataya2023will, sariyildiz2023fake}.
However, in the context of noisy supervised learning, these limitations are outweighed by a more critical need: establishing reliable and stable anchors for each category to mitigate information loss in noisy datasets. 
In this scenario, synthetic images emerge as a particularly valuable resource, as they naturally maintain high semantic consistency between their content and labels - a property that directly addresses the core challenge of semantic noise in real-world datasets.

Our key insight is that while synthetic images may not fully capture the complexity and diversity of real-world data distributions, they can serve as reliable reference points for identifying and correcting mislabeled samples. By leveraging the semantic consistency of synthetic images rather than treating them as direct training data, we can circumvent their limitations while exploiting their inherent advantages in label reliability.
In this paper, we propose a simple yet effective method that leverages generated images to re-label noisy training data, thus supporting robust model training under noisy supervision. 
Through extensive experiments across multiple datasets and noise settings, we demonstrate that our synthetic image-based relabeling approach can effectively filter and refine noisy labels, significantly improving model accuracy under standard training procedures. 
Our method operates as a preprocessing step before model training, making it orthogonal to and readily integrable with existing noisy label learning approaches. This complementary nature enables synergistic combinations with state-of-the-art techniques to achieve superior performance across various noise conditions.

Our contributions are as follows:
\begin{itemize}[itemsep=0pt, topsep=2pt]
    \item We first validate the effectiveness of synthetic images for noisy label learning, providing valuable insights for future research in this field.
    \item Our method can be directly integrated with other noisy label learning approaches, resulting in higher accuracy and establishing state-of-the-art performance.
\end{itemize}

\section{Related work}
\subsection{Noisy label learning}
\label{sec:related_noisy}

Conventional research on noisy label learning~\cite{tanaka2018joint, mao2021noisy} was primarily based on an i.i.d. assumption, i.e., the corruption of labels is independent and identically distributed. 
However, this assumption does not align with real-world scenarios, as noise in real datasets tends to be feature-dependent. 
Zhang et al.~\cite{zhang2021learning} introduced a feature-dependent label noise called Polynomial Margin Diminishing (PMD) label noise and proposed a progressive label correction algorithm.
Chen et al.~\cite{chen2024label} employed pre-trained encoders to retrieve pseudo-clean labels, which were then used by diffusion models to refine noisy labels. Their method achieved state-of-the-art results on various noisy label benchmarks. 
We further enhance these outcomes by utilizing synthetic images to generate more reliable pseudo-clean labels.

\subsection{Synthetic images for image classification}
\label{sec:related_synthetic}

In recent years, the advancement of text-to-image diffusion models has led to studies utilizing images generated by these models for image classification tasks~\cite{hataya2023will, sariyildiz2023fake, he2022synthetic}.
Hataya et al.~\cite{hataya2023will} created ImageNet-SD using Stable Diffusion models to mirror ImageNet-1K~\cite{russakovsky2015imagenet}. 
They replaced real images in the ImageNet training set with synthetic images and found that increasing the proportion of synthetic images reduced classification accuracies. 
Sariyildiz et al.~\cite{sariyildiz2023fake} improved accuracy by tuning prompts for synthetic image generation, though results still fell short of those with real images. 
These studies indicate that incorporating synthetic images into training sets can reduce accuracy, mainly because of their limited diversity and inability to capture complex real-world scenarios.
In this paper, we do not use synthetic images directly for training; instead, we enhance the semantic consistency of synthetic images, making them reliable and stable anchors for image classification tasks.

\section{Approach}
\label{sec:approach}

\begin{figure*}[t]
    \centering
    \includegraphics[width=\linewidth]{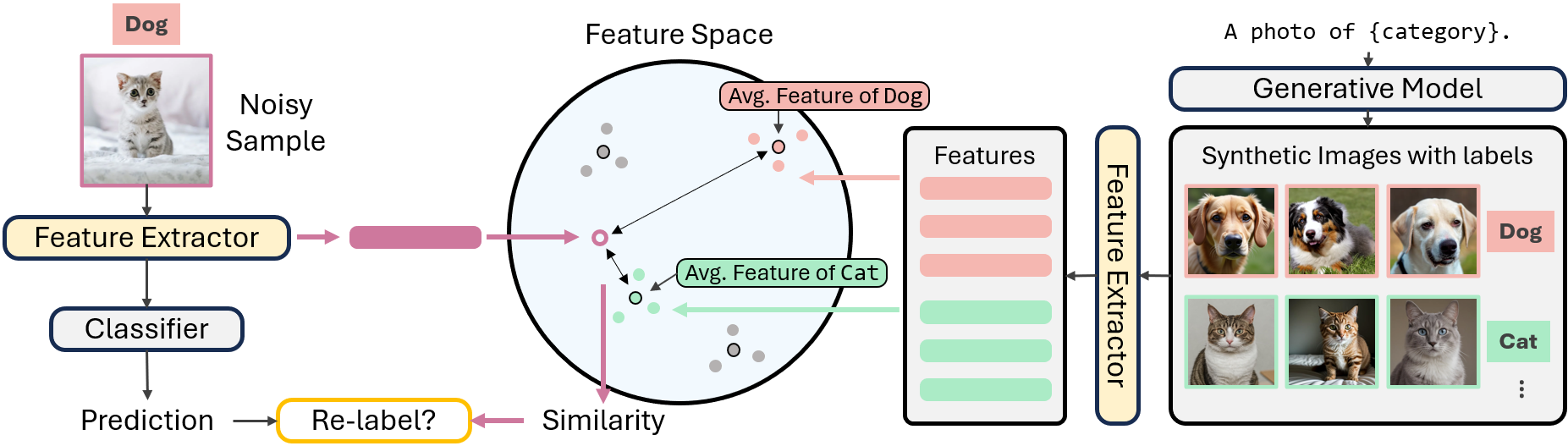}
    \caption{Given a noisy sample, a feature extractor and classifier first process the image and make an initial prediction. Meanwhile, synthetic images generated by a generative model provide reliable reference points in feature space. The final re-labeling decision leverages both the classifier's prediction and the sample's feature similarity to synthetic class representations}
    \label{fig:pipeline}
\end{figure*}


Our proposed method addresses the challenge of learning from a noisy dataset comprising $M$ real images $\mathcal{R} = \{r_i\}_{i=1}^M$ with potentially incorrect labels $\mathcal{L} = \{l_i\}_{i=1}^M$, where $l_i \in \mathcal{C}$ belongs to a predefined set of categories $\mathcal{C}$, e.g., $\mathcal{C} = \{''\texttt{dog}'', ''\texttt{cat}'', \dots \}$. 
As illustrated in Fig.~\ref{fig:pipeline}, the core idea of our approach is to take advantage of synthetic images as reliable references to refine these noisy labels into a corrected set $\mathcal{L}'$. 
Our approach first uses standard training on the noisy dataset to obtain a classifier $\{\mathcal{E}, f\}$, consisting of a feature extractor $\mathcal{E}$ and a classification layer $f$. 
By generating synthetic images and using $\mathcal{E}$ to compute their class prototypes, we create reliable reference features for each category. The final label correction decision for samples is made according to both the classifier's confidence scores and the feature similarities with these synthetic prototypes. 
This process is outlined in Algorithm~\ref{alg:algorithm}. In the following, we explain each step in detail.


\noindent\textbf{\textit{Synthetic reference generation.}}
We generate images for each category $c \in \mathcal{C}$ using a text-to-image model with the simple prompt ``\texttt{A photo of $c$}'', resulting in $N$ synthetic images $\mathcal{G} = \{g_j\}_{j=1}^N$ in total. These images inherently carry reliable labels $c$ given by their generating prompts. 

\noindent\textbf{\textit{Feature extractor.}} 
We train a $|\mathcal{C}|$-class classifier $\{\mathcal{E}, f\}$ on the noisy dataset $(\mathcal{R}, \mathcal{L})$. While this classifier may not achieve optimal performance due to label noise, we observe that learned feature representation of $\mathcal{E}$ remains informative for similarity-based label correction. 

\noindent\textbf{\textit{Synthetic prototype construction.}}
For each category $c$, we construct a reliable prototype $\mathbf{p}_c$ by averaging the features of its corresponding synthetic images. These prototypes serve as anchor points in feature space.

\noindent\textbf{\textit{Re-labeling.}} 
We propose a dual-criteria approach for label correction that leverages both feature similarity and classifier confidence given by $\{\mathcal{E}, f\}$ to update $\mathcal{L}$. For each real image $r$, we: $(1)$ Calculate cosine similarities $S^\text{sim}$ between its features and all synthetic prototypes. $(2)$ Obtain classification confidence scores $S^\text{conf}$ from the trained classifier $\{\mathcal{E}, f\}$. $(3)$ Compute a combined score for each category $c$ as follows,
\begin{equation}
    S_c = \alpha \cdot S^\text{sim}_c + (1-\alpha) \cdot S^\text{conf}_c
\end{equation}
where $\alpha$ is constant, balancing the contribution of similarity and classification confidence. We update the label to the category with the highest score only if this score exceeds a threshold $\theta$, otherwise maintaining the original label. This procedure yields a refined label set $\mathcal{L}'$.

The resulting refined dataset $\{\mathcal{R}, \mathcal{L}'\}$ can be used either for standard training or as input to existing noise-robust learning methods such as \textit{PLC}~\cite{zhang2021learning} and \textit{LRA-diffusion}~\cite{chen2024label}, demonstrating significant performance improvements in both scenarios.

\begin{algorithm}[t!]
\caption{Noisy label refinement using synthetic images.}
\label{alg:algorithm}
\begin{minipage}{\linewidth}
    \textbf{Input}: Real images $\mathcal{R}$ with labels $\mathcal{L}$ \\
    \textbf{Parameter}: Threshold $\theta$ \\
    \textbf{Output}: Re-labeled real images' labels $\mathcal{L}'$
\end{minipage}
\begin{algorithmic}[1] 
    \STATE Generate images and train a classifier: \\
    $\mathcal{G} \gets $ Generated using $\mathcal{L}$ \\
    $\mathcal{E}, f \gets$ Trained by $\mathcal{R}$ and $\mathcal{L}$\\
    \STATE Calculate prototypes for each category of $\mathcal{G}$: \\
    $\mathcal{G}_c \gets \{ g \in \mathcal{G} \mid l_g = c\}, ~~~ \forall c \in \mathcal{C}$ \\
    $\mathbf{p}_c \gets \frac{1}{|\mathcal{G}_c|}\sum_{g \in \mathcal{G}_c} \mathcal{E}(g), ~~~ \forall c \in \mathcal{C}$
    \FOR{$r \in \mathcal{R}$ and its corresponding label $l \in \mathcal{L}$}
        \STATE Compute cosine similarities for all categories: \\
        $S^\text{sim}_c \gets \text{CosSim}(\mathcal{E}(r), \mathbf{p}_c), ~~~ \forall c \in \mathcal{C}$
        \STATE Compute classification accuracies for all categories: \\
        $S^\text{conf}_c \gets f(r)[c], ~~~ \forall c \in \mathcal{C}$
        \STATE Calculate scores for all categories: \\
        $S_c \gets \alpha \cdot S^\text{sim}_c + (1-\alpha) \cdot S^\text{conf}_c, ~~~ \forall c \in \mathcal{C}$
        \STATE Determine the final score and candidate label: \\
        $s_\text{new} \gets \max(\mathbf{S})$ \\
        $l_\text{new} \gets \argmax_{c}(\mathbf{S})$
        \IF{$s_\text{new} \geq \theta$}
            \STATE $l' \gets l_\text{new}$
        \ELSE
            \STATE $l' \gets l$
        \ENDIF
        \STATE Append $l'$ to $\mathcal{L}'$
    \ENDFOR
    \STATE \textbf{return} $\mathcal{L}'$
\end{algorithmic}
\end{algorithm}

\section{Experiments}
\subsection{Experimental setup}
\label{sec:settings}

\begin{table*}[t]
\centering
\caption{Classification accuracies (\%) on CIFAR-10 and CIFAR-100 under feature-dependent noise and hybrid noise, combining PMD noise with Uniform (U) and Asymmetric (A) noise.}
\label{tab:main_results}

\begin{subtable}[t]{\textwidth}
\centering
\begin{tabular}{@{}llllll@{}}
\toprule
 & \multicolumn{5}{c}{\textbf{CIFAR-10}} \\ 
\cmidrule(l){2-6}
\textbf{Methods} & \small{35\% PMD} & \small{70\% PMD} & \small{35\% PMD + 30\% U} & \small{35\% PMD + 60\% U} & \small{35\% PMD + 30\% A} \\
\midrule
\small{Standard} & 80.81 & 40.32 & 77.12 & 67.92 & 77.40 \\
\small{Re-labeled data (Ours)} & 82.46 \scalebox{0.7}{(+1.65)} & 54.10 \scalebox{0.7}{(+13.78)} & 79.76 \scalebox{0.7}{(+2.64)} & 72.38 \scalebox{0.7}{(+4.46)} & 78.97 \scalebox{0.7}{(+1.57)} \\
\midrule
\small{PLC} & 82.87 & 38.65 & 77.98 & 61.87 & 78.52 \\
\small{PLC + Ours} & 83.26 \scalebox{0.7}{(+0.39)} & 54.01 \scalebox{0.7}{(+15.36)} & 81.38 \scalebox{0.7}{(+3.40)} & 72.57 \scalebox{0.7}{(+10.70)} & 80.49 \scalebox{0.7}{(+1.97)} \\
\midrule
\small{LRA-diffusion (SimCLR)} & 89.29 & 40.87 & 88.83 & 83.67 & 86.89 \\
\small{LRA-diffusion (SimCLR) + Ours} & 88.87 \scalebox{0.7}{(-0.42)} & 60.78 \scalebox{0.7}{(+19.91)} & 88.41 \scalebox{0.7}{(-0.42)} & 85.31 \scalebox{0.7}{(+1.64)} & 87.95 \scalebox{0.7}{(+1.06)} \\
\small{LRA-diffusion (CLIP)} & 96.91 & 41.78 & 96.61 & 88.66 & 94.35 \\
\small{LRA-diffusion (CLIP) + Ours} & \textbf{96.93} \scalebox{0.7}{(+0.02)} & \textbf{71.59} \scalebox{0.7}{(+29.81)} & \textbf{96.68} \scalebox{0.7}{(+0.07)} & \textbf{94.94} \scalebox{0.7}{(+6.28)} & \textbf{96.78} \scalebox{0.7}{(+2.43)} \\
\bottomrule
\end{tabular}
\end{subtable}

\vspace{8pt}

\begin{subtable}[t]{\textwidth}
\centering
\begin{tabular}{@{}llllll@{}}
\toprule
 & \multicolumn{5}{c}{\textbf{CIFAR-100}} \\ 
\cmidrule(l){2-6}
\textbf{Methods} & \small{35\% PMD} & \small{70\% PMD} & \small{35\% PMD + 30\% U} & \small{35\% PMD + 60\% U} & \small{35\% PMD + 30\% A} \\
\midrule
\small{Standard} & 59.28 & 44.43 & 55.98 & 43.50 & 52.23 \\
\small{Re-labeled data (Ours)} & 61.45 \scalebox{0.7}{(+2.17)} & 47.15 \scalebox{0.7}{(+2.72)} & 59.35 \scalebox{0.7}{(+3.37)} & 44.91 \scalebox{0.7}{(+1.41)} & 61.39 \scalebox{0.7}{(+9.16)} \\
\midrule
\small{PLC} & 60.06 & 45.03 & 57.67 & 38.92 & 59.34 \\
\small{PLC + Ours} & 59.59 \scalebox{0.7}{(-0.47)} & 46.13 \scalebox{0.7}{(+1.10)} & 59.43 \scalebox{0.7}{(+1.76)} & 43.25 \scalebox{0.7}{(+4.33)} & 60.08 \scalebox{0.7}{(+0.74)} \\
\midrule
\small{LRA-diffusion (SimCLR)} & 54.95 & 48.00 & 55.23 & 47.47 & 53.99 \\
\small{LRA-diffusion (SimCLR) + Ours} & 56.18 \scalebox{0.7}{(+1.23)} & 49.34 \scalebox{0.7}{(+1.34)} & 54.67 \scalebox{0.7}{(-0.56)} & 48.54 \scalebox{0.7}{(+1.07)} & 55.45 \scalebox{0.7}{(+1.46)} \\
\small{LRA-diffusion (CLIP)} & 75.91 & 56.15 & 74.69 & 63.08 & 69.97 \\
\small{LRA-diffusion (CLIP) + Ours} & \textbf{77.01} \scalebox{0.7}{(+1.10)} & \textbf{66.88} \scalebox{0.7}{(+10.73)} & \textbf{76.10} \scalebox{0.7}{(+1.41)} & \textbf{66.27} \scalebox{0.7}{(+3.19)} & \textbf{74.42} \scalebox{0.7}{(+4.45)} \\
\bottomrule
\end{tabular}
\end{subtable}

\end{table*}

\noindent\textbf{\textit{Datasets and noise types.}}
Following existing noisy label learning methods~\cite{zhang2021learning, chen2024label}, we conduct experiments on the CIFAR-10 and CIFAR-100 datasets~\cite{krizhevsky2009learning} under various noise types. 
We test our method with two types of noise: \textbf{feature-dependent noise} and \textbf{hybrid noise}.
For feature-dependent noise, we utilize the polynomial margin diminishing (PMD) noise~\cite{zhang2021learning}, applying it at rates of 35\% and 70\%. 
For hybrid noise, we combine PMD noise with two types of independent and identically distributed (i.i.d.) noise: uniform noise, where the noisy labels are evenly distributed among all remaining classes; 
and asymmetric noise, where labels are mislabeled as only one specific class. 

Furthermore, to validate the generalizability of our proposed method, we also conducted experiments on the ImageNet-100 dataset, which is a subset of ImageNet-1k~\cite{russakovsky2015imagenet}.

\noindent\textbf{\textit{Synthetic images.}}
We employ SDXL-Turbo~\cite{sauer2025adversarial} as our generative model, which is a distilled version of Stable Diffusion XL~\cite{podell2023sdxl} generating images of size $512 \times 512$. 
We generate 100 images for each category $c \in \mathcal{C}$.

\noindent\textbf{\textit{Implementation details.}}
We utilize ResNet34 for CIFAR-10 and CIFAR-100 datasets, and ResNet50 for ImageNet-100 datasets~\cite{he2016deep}. 
During the preprocessing phase, we train classifiers for 200 epochs using the SGD optimizer. The batch size is 128 and the initial learning rate is set at 0.1. 
After obtaining the re-labeled data, we fine-tune the classifiers for 50 epochs with an learning rate of 0.001. 
All experiments are conducted on an NVIDIA A100 GPU. 
Generating 100 images per category takes 3-4 minutes, and re-labeling the entire CIFAR dataset takes about 40 minutes in total---far less than the 8+ hours typically required by robust training methods.

\subsection{Main results}
\label{sec:main_exp}

Table~\ref{tab:main_results} shows the results on CIFAR-10 and CIFAR-100.
\textit{Standard} refers to the accuracy of classifiers trained on noisy data. \textit{Re-labeled data} represents our proposed method, which involves using synthetic images to refine noisy labels and then fine-tuning the classifier with the updated labels. Based on our experiments, we set the threshold at 0.6.
As shown in Table~\ref{tab:main_results}, our method improves classification accuracy.

Additionally, we combined the re-labeled data with existing noisy label learning methods to further validate the effectiveness of our proposed approach. 
Although there are many challenging methods available~\cite{zhang2018generalized, wang2019symmetric, zheng2020error, zhao2022centrality}, we selected two of the most advanced and effective: the \textit{PLC} method~\cite{zhang2021learning} and the \textit{LRA-diffusion} method~\cite{chen2024label}. 
For the \textit{LRA-diffusion} method, we used two pre-trained encoders: (1) SimCLR, provided by~\cite{chen2024label}, which is a ResNet50 trained on the CIFAR-10 and CIFAR-100 datasets through contrastive learning; (2) CLIP~\cite{radford2021learning}, following~\cite{chen2024label}, we used the vision transformer encoder (ViT-L/14)~\cite{dosovitskiy2020image} with pre-trained weights.
The results show that combining our method with existing methods can further enhance accuracy. When integrated with the current state-of-the-art method, \textit{LRA-diffusion (CLIP)}, our approach achieves even higher accuracy.
Notably, in challenging noise conditions, such as 70\% PMD noise, combining our method with existing approaches results in a significant improvement of 29.81\% on CIFAR-10 and 10.73\% on CIFAR-100. This indicates that in complex real-world scenarios, synthetic images can serve as a reliable anchor.

\subsection{Ablation study}
\label{sec:ablation}

\noindent\textbf{\textit{Domain adaptation.}}
Due to the domain gap between images generated by the Stable Diffusion model and those from the CIFAR-10 and CIFAR-100 datasets, which might affect the results, we fine-tuned the SDXL-Turbo model with noisy data of CIFAR-100. 
The images generated by the fine-tuned model more closely match CIFAR-100 in appearance and feature distribution. 
However, incorporating these synthetic images into our noisy label learning did not significantly improve accuracy. 
Table~\ref{tab:finetune} shows the classification accuracies on CIFAR-100 with 70\% PMD noise, indicating that the impact of fine-tuning the Stable Diffusion model was minimal. 
Given the extra time and computational resources required, we chose not to fine-tune the Stable Diffusion model in our approach.

\begin{table}[t]
\centering
\caption{Impact of fine-tuning Stable Diffusion models on classification accuracies (\%).}
\label{tab:finetune}
\begin{tabular}{@{}lcc@{}}
\toprule
Methods & W/o tuning & W/ tuning \\
\midrule
\small{Re-labeled data (Ours)} & 47.15 & 47.40 \\
\small{LRA-diffusion (SimCLR) + Ours} & 49.34 & 50.57 \\
\small{LRA-diffusion (CLIP) + Ours} & 66.88 & 66.76 \\
\bottomrule
\end{tabular}
\end{table}

\begin{table}[t]
\centering
\caption{Impact of feature extractor changes on results.}
\label{tab:extrator}
\begin{tabular}{@{}lc@{}}
\toprule
Feature extractor & Accuracy (\%) \\
\midrule
CLIP & 40.74 \\
ResNet34 trained on synthetic images & 44.08 \\
ResNet34 trained on noisy labels & \textbf{46.06} \\
\midrule
ResNet34 trained on clean labels & 52.64 \\
\bottomrule
\end{tabular}
\end{table}

\noindent\textbf{\textit{Feature extractor.}}
We evaluated the impact of using different feature extractors on calculating the similarities between real and synthetic images, and the classification accuracies on CIFAR-100 with 70\% PMD noise are shown in Table~\ref{tab:extrator}.
We employed CLIP, a ResNet34 trained on synthetic images, and a ResNet34 trained on noisy labels as feature extractors. 
To establish an upper reference bound, we also conducted experiments using a ResNet34 trained on clean labels. 
Since this experiment aimed to compare feature extraction capabilities, we set $S_c = S^\text{sim}_c$ to avoid possible interferences.
As shown in Table~\ref{tab:extrator}, even the classifier trained on data with 70\% PMD noise can effectively extract features from synthetic images, thus providing a reliable anchor for noisy label refinement.

\noindent\textbf{\textit{Parameters.}}
Table~\ref{tab:parameter} shows how varying the parameter $\alpha$ in $S_c = \alpha \cdot S^\text{sim}_c + (1-\alpha) \cdot S^\text{conf}_c$, as well as different thresholds $\theta$, affect classification accuracies on the CIFAR-100 dataset, which is used as training data with 70\% PMD noise.
$S^\text{sim}_c$ represents the similarities between real and synthetic images, while $S^\text{conf}_c$ represents the classifier’s predictions of the probability distribution across categories. The threshold indicates the extent to which the original label is considered.
Our experiments demonstrate that by considering these three elements together, we can achieve the highest quality of re-labeled data and the best fine-tuned classification accuracy.
We have identified the optimal parameters ($\alpha=0.5, \theta=0.6$) through extensive testing.

\begin{table}[t]
\centering
\caption{Impact of parameters on accuracies (\%).}
\label{tab:parameter}
\begin{tabular}{@{}ccccc@{}}
\toprule
Threshold $\theta$ & $\alpha=1$ & $\alpha=0.7$ & $\alpha=0.5$ & $\alpha=0.3$ \\
\midrule
0 & 31.14 & 42.41 & 43.69 & 43.84 \\
0.6 & 46.06 & 47.10 & \textbf{47.40} & 47.19 \\
\bottomrule
\end{tabular}
\end{table}

\subsection{Results on ImageNet datasets}
\label{sec:ImageNet}

To validate the generalizability of our proposed method, we conducted experiments on the ImageNet-100 dataset. 
We tested our method on training sets with Uniform (U) noise and Asymmetric (A) noise, and the experimental results are shown in Table~\ref{tab:ImageNet}.
The results demonstrate that our proposed method is effective, particularly in more complex noise scenarios, such as 60\% asymmetric noise. This further validates the insights obtained from our main results.


\begin{table}[t]
\centering
\caption{Classification accuracies (\%) on ImageNet-100.}
\label{tab:ImageNet}
\begin{tabular}{@{}lcccc@{}}
\toprule
Methods & 30\% U & 60\% U & 30\% A & 60\% A \\
\midrule
\small{Standard} & 66.54 & 49.60 & 74.22 & 20.74 \\
\small{Re-labeled data (Ours)} & 69.26 & 52.36 & 78.28 & 44.84 \\
\bottomrule
\end{tabular}
\end{table}

\section{Conclusion}
\label{sec:conclusion}

In this paper, we address the challenge of learning from noisy labels by leveraging the semantic reliability of synthetic images. 
We utilize these images as reliable and stable anchors for refining noisy labels, and our experiments across various datasets demonstrate that our proposed method significantly improves classification accuracies, particularly under challenging noise conditions.
Furthermore, as our approach is orthogonal to existing noisy label learning methods, it can be seamlessly integrated with state-of-the-art techniques, thereby setting a new benchmark in the field. 
Our work not only highlights the potential of synthetic images to enhance supervised learning in noisy environments but also offers significant insights for the community, encouraging further exploration into the use of synthetic data for more robust machine learning applications.

\bibliographystyle{IEEEbib}
\bibliography{strings,refs}

\begin{thebibliography}{10}

\bibitem{zhang2021learning}
Yikai Zhang, Songzhu Zheng, Pengxiang Wu, Mayank Goswami, and Chao Chen,
\newblock ``Learning with feature-dependent label noise: A progressive approach,''
\newblock in {\em Proceedings of the International Conference on Learning Representations}, 2021.

\bibitem{ramesh2022hierarchical}
Aditya Ramesh, Prafulla Dhariwal, Alex Nichol, Casey Chu, and Mark Chen,
\newblock ``Hierarchical text-conditional image generation with clip latents,''
\newblock {\em arXiv preprint arXiv:2204.06125}, vol. 1, no. 2, pp. 3, 2022.

\bibitem{rombach2022high}
Robin Rombach, Andreas Blattmann, Dominik Lorenz, Patrick Esser, and Bj{\"o}rn Ommer,
\newblock ``High-resolution image synthesis with latent diffusion models,''
\newblock in {\em Proceedings of the IEEE/CVF conference on computer vision and pattern recognition}, 2022, pp. 10684--10695.

\bibitem{hataya2023will}
Ryuichiro Hataya, Han Bao, and Hiromi Arai,
\newblock ``Will large-scale generative models corrupt future datasets?,''
\newblock in {\em Proceedings of the IEEE/CVF International Conference on Computer Vision}, 2023, pp. 20555--20565.

\bibitem{sariyildiz2023fake}
Mert~Bulent Sariyildiz, Karteek Alahari, Diane Larlus, and Yannis Kalantidis,
\newblock ``Fake it till you make it: Learning transferable representations from synthetic imagenet clones,''
\newblock in {\em Proceedings of the IEEE/CVF Conference on Computer Vision and Pattern Recognition}, 2023.

\bibitem{tanaka2018joint}
Daiki Tanaka, Daiki Ikami, Toshihiko Yamasaki, and Kiyoharu Aizawa,
\newblock ``Joint optimization framework for learning with noisy labels,''
\newblock in {\em Proceedings of the IEEE Conference on Computer Vision and Pattern Recognition}, 2018, pp. 5552--5560.

\bibitem{mao2021noisy}
Jiafeng Mao, Qing Yu, Yoko Yamakata, and Kiyoharu Aizawa,
\newblock ``Noisy annotation refinement for object detection,''
\newblock {\em arXiv preprint arXiv:2110.10456}, 2021.

\bibitem{chen2024label}
Jian Chen, Ruiyi Zhang, Tong Yu, Rohan Sharma, Zhiqiang Xu, Tong Sun, and Changyou Chen,
\newblock ``Label-retrieval-augmented diffusion models for learning from noisy labels,''
\newblock {\em Advances in Neural Information Processing Systems}, vol. 36, 2024.

\bibitem{he2022synthetic}
Ruifei He, Shuyang Sun, Xin Yu, Chuhui Xue, Wenqing Zhang, Philip Torr, Song Bai, and Xiaojuan Qi,
\newblock ``Is synthetic data from generative models ready for image recognition?,''
\newblock {\em arXiv preprint arXiv:2210.07574}, 2022.

\bibitem{russakovsky2015imagenet}
Olga Russakovsky, Jia Deng, Hao Su, Jonathan Krause, Sanjeev Satheesh, Sean Ma, Zhiheng Huang, Andrej Karpathy, Aditya Khosla, Michael Bernstein, et~al.,
\newblock ``Imagenet large scale visual recognition challenge,''
\newblock {\em International Journal of Computer Vision}, vol. 115, pp. 211--252, 2015.

\bibitem{krizhevsky2009learning}
Alex Krizhevsky, Geoffrey Hinton, et~al.,
\newblock ``Learning multiple layers of features from tiny images,''
\newblock 2009.

\bibitem{sauer2025adversarial}
Axel Sauer, Dominik Lorenz, Andreas Blattmann, and Robin Rombach,
\newblock ``Adversarial diffusion distillation,''
\newblock in {\em Proceedings of the European Conference on Computer Vision}. Springer, 2025, pp. 87--103.

\bibitem{podell2023sdxl}
Dustin Podell, Zion English, Kyle Lacey, Andreas Blattmann, Tim Dockhorn, Jonas M{\"u}ller, Joe Penna, and Robin Rombach,
\newblock ``Sdxl: Improving latent diffusion models for high-resolution image synthesis,''
\newblock {\em arXiv preprint arXiv:2307.01952}, 2023.

\bibitem{he2016deep}
Kaiming He, Xiangyu Zhang, Shaoqing Ren, and Jian Sun,
\newblock ``Deep residual learning for image recognition,''
\newblock in {\em Proceedings of the IEEE Conference on Computer Vision and Pattern Recognition}, 2016, pp. 770--778.

\bibitem{zhang2018generalized}
Zhilu Zhang and Mert Sabuncu,
\newblock ``Generalized cross entropy loss for training deep neural networks with noisy labels,''
\newblock {\em Advances in Neural Information Processing Systems}, vol. 31, 2018.

\bibitem{wang2019symmetric}
Yisen Wang, Xingjun Ma, Zaiyi Chen, Yuan Luo, Jinfeng Yi, and James Bailey,
\newblock ``Symmetric cross entropy for robust learning with noisy labels,''
\newblock in {\em Proceedings of the IEEE/CVF International Conference on Computer Vision}, 2019, pp. 322--330.

\bibitem{zheng2020error}
Songzhu Zheng, Pengxiang Wu, Aman Goswami, Mayank Goswami, Dimitris Metaxas, and Chao Chen,
\newblock ``Error-bounded correction of noisy labels,''
\newblock in {\em Proceedings of the International Conference on Machine Learning}. PMLR, 2020, pp. 11447--11457.

\bibitem{zhao2022centrality}
Ganlong Zhao, Guanbin Li, Yipeng Qin, Feng Liu, and Yizhou Yu,
\newblock ``Centrality and consistency: two-stage clean samples identification for learning with instance-dependent noisy labels,''
\newblock in {\em Proceedings of the European Conference on Computer Vision}. Springer, 2022, pp. 21--37.

\bibitem{radford2021learning}
Alec Radford, Jong~Wook Kim, Chris Hallacy, Aditya Ramesh, Gabriel Goh, Sandhini Agarwal, Girish Sastry, Amanda Askell, Pamela Mishkin, Jack Clark, et~al.,
\newblock ``Learning transferable visual models from natural language supervision,''
\newblock in {\em Proceedings of the International Conference on Machine Learning}. PMLR, 2021, pp. 8748--8763.

\bibitem{dosovitskiy2020image}
Alexey Dosovitskiy, Lucas Beyer, Alexander Kolesnikov, Dirk Weissenborn, Xiaohua Zhai, Thomas Unterthiner, Mostafa Dehghani, Matthias Minderer, Georg Heigold, Sylvain Gelly, et~al.,
\newblock ``An image is worth 16x16 words: Transformers for image recognition at scale,''
\newblock {\em arXiv preprint arXiv:2010.11929}, 2020.

\end{thebibliography}

\end{document}